\documentclass{article}
\pdfoutput=1
\usepackage[backend=biber]{biblatex}
\usepackage{spconf,amsmath,graphicx}

\usepackage{times}
\usepackage{latexsym}

\usepackage[T1]{fontenc}
\usepackage[utf8]{inputenc}
\usepackage{microtype}
\usepackage{color}
\usepackage{subfigure}
\usepackage{times}
\usepackage{latexsym}
\usepackage{amssymb}
\usepackage{amsmath}
\usepackage{tabularx} 
\usepackage{array}
\usepackage{booktabs} 
\usepackage{amsthm}
\usepackage{multirow}
\usepackage{float}
\usepackage{enumitem}
\usepackage{algorithm, algorithmic}

\usepackage{graphicx}
\usepackage{bbm}

\title{Mutually Guided Few-shot Learning for Relational Triple Extraction}
\name{Chengmei Yang$^{\star \dagger}$ \qquad Shuai Jiang $^{\star} $ \qquad Bowei He$^{\dagger}$ \qquad Chen Ma$^{\dagger}$ \qquad Lianghua He$^{\star}$}

\address{$^{\star}$ Department of Computer Science and Technology, Tongji University, China \\
$^{\dagger}$ Department of Computer Science, City University of Hong Kong, Hong Kong \\
\{ycm, Helianghua, 2032980\}@tongji.edu.cn, boweihe2-c@my.cityu.edu.hk, chenma@cityu.edu.hk}

\begin{document}
%
\maketitle
\begin{abstract}
Knowledge graphs (KGs), containing many entity-relation-entity triples, provide rich information for downstream applications. Although extracting triples from unstructured texts has been widely explored, most of them require a large number of labeled instances. The performance will drop dramatically when only few labeled data are available.
To tackle this problem, we propose the \textbf{M}utually \textbf{G}uided \textbf{F}ew-shot learning framework for Relational \textbf{T}riple \textbf{E}xtraction (MG-FTE). Specifically, our method consists of an \textit{entity-guided relation proto-decoder} to classify the relations firstly and a \textit{relation-guided entity proto-decoder} to extract entities based on the classified relations. To draw the connection between entity and relation, we design a \textit{proto-level fusion module} to boost the performance of both entity extraction and relation classification.
Moreover, a new cross-domain few-shot triple extraction task is introduced. Extensive experiments show that our method outperforms many state-of-the-art methods by $12.6$ F1 score on FewRel 1.0 (single-domain) and $20.5$ F1 score on FewRel 2.0 (cross-domain). \renewcommand{\thefootnote}{\roman{footnote}}
\footnotetext{Lianghua He and Chen Ma are both corresponding authors. Our source code is publicly available at https://github.com/ycm094/MG-FTE-main.} 
\end{abstract}
\begin{keywords}
Few-shot Triple Extraction, Information Extraction, Knowledge Graph
\end{keywords}
\section{Introduction}
Thanks to its effectiveness for modeling relational data, the knowledge graph (KG) has been deployed in many applications \cite{recommendation1,recommendation2}. A KG consists of a quantity of entity-relation-entity triples, which illustrates a certain relation between two entities, such as (\textit{Tom Hanks, starred\_in, Forrest Gump}). 

To make the KG fulfill its capacity, the construction of triples plays a fundamental role. The conventional triple extraction task aims at extracting entities and relations simultaneously in a unified form of (\emph{subject, relation, object}), from unstructured texts through supervised learning methods \cite{CasRel, ICASSP22}. Though effective, these approaches usually need 
\emph{a large number of labeled instances} to bridge the potential relationship between sentences and relational triples to be extracted. 
Whereas, due to the long-tail distribution of relations \cite{FewRel1, DBLP:conf/mm/ZhengFFCL021}, it is worth considering the relational triple extraction with insufficient training instances.
Consequently, to enable typical relational triple extraction approaches in the few-shot setting, few-shot learning techniques \cite{DBLP:journals/pami/Fei-FeiFP06, prototype} are leveraged to learn how to connect texts and triples with only \emph{a few labeled instances}. 

However, simply combining traditional triple extraction approaches with few-shot learning methods is far from satisfactory.
Firstly, \textbf{the discrepancy of subjects/objects under different relations is large in the few-shot setting}. Previous work \cite{COLING20} recognizes entities at first and then classifies relations based on extracted entities. We argue that if the entity is extracted from sentences without considering the distinction of relations, the prototypes of its subjects and objects are difficult to be representative. That is, the model is more difficult to learn the prototypes that can be compatible with the representation of subjects and objects under all relations. 
Secondly, \textbf{plenty of non-relational entities 
and words may mislead the relation extraction.} A relation/semantic prototype is required to consider all word representation in one sentence. Whereas, a triple only consists of two relational entities, which means there are plenty of non-relational entities and words that make no contribution to the certain relation. Therefore, learning a representative relation/semantic prototype is crucial.
Thirdly, \textbf{the connection between entities and relations/semantics is hard to learn in the few-shot setting}. Conventional methods that jointly extract triples \cite{ICASSP22, CasRel} are difficult to construct an optimal matching between relational entities and relations, 
since the entity and relation prototypes for each few-shot task are only based on a few provided instances, which makes the learned prototypes incomplete \cite{cvpr21}.


To tackle the aforementioned challenges, we propose a mutually guided framework consisting of an entity-guided relation proto-decoder (EGr) and a relation-guided entity proto-decoder (RGe).
Moreover, to strengthen the connection between entities and relations/semantics, both relation and entity proto-decoders are built upon our proposed proto-level fusion module.
We also introduce this task into a more challenging and practical scenario---cross-domain few-shot relational triple extraction, which contributes to many real-world applications \cite {cross-domain-application1, cross-domain-application2}. For example, when constructing a knowledge graph for a special field, like tourism, medical or finance, it is time-consuming and labor-intensive to ask professionals for annotation. However, there are plenty of well-annotated and easy-available universal knowledge graph datasets, such as FreeBase \cite{Freebase} and DBpedia \cite{DBpedia}, which can be utilized for training a model and then generalizing it to a certain domain. Therefore, with only a few labeled instances in the target domain, the model can automatically extract triples from texts to facilitate the specific knowledge graph construction.
To summarize, our contributions are: 1) We propose a novel mutually guided framework, consisting of an entity-guided relation proto-decoder to classify relations at first and a relation-guided entity proto-decoder to extract entities based on classified relations.
2) We design a proto-level fusion module to strengthen the connection between relational entities and relation/semantics, which benefits both of our mutually guided proto-decoders. 
3) We first engage the few-shot triple extraction task into the cross-domain setting and the experimental results under both single-domain and cross-domain settings demonstrate the effectiveness of our method.

\begin{figure*}[h]
 \centering
 \includegraphics[width=0.95\linewidth]{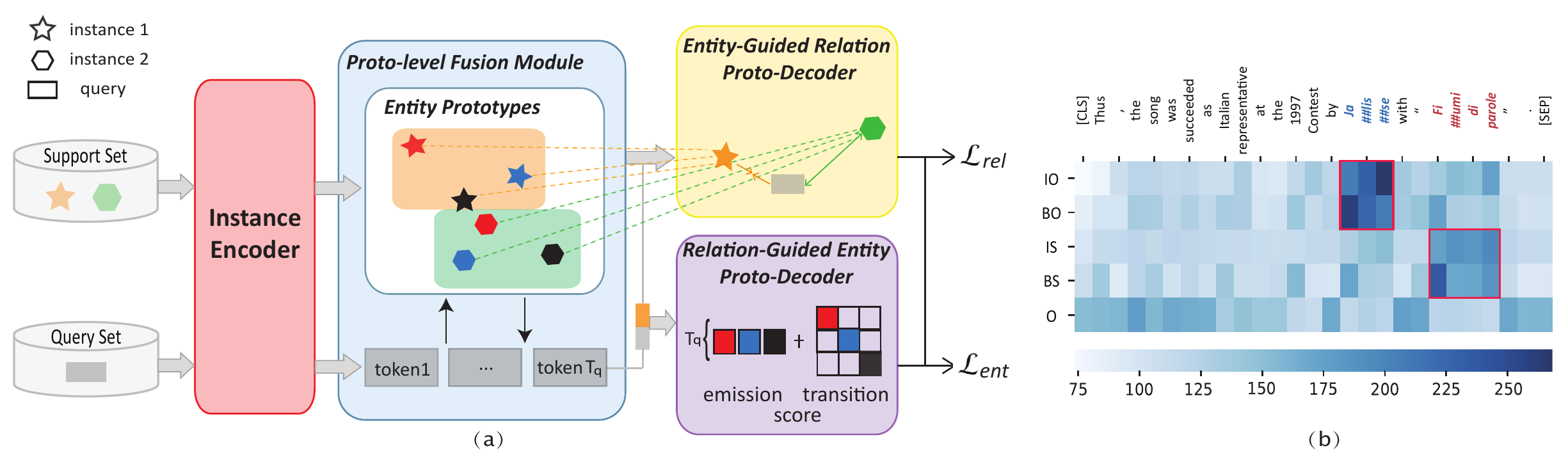}
 \caption{Subfigure (a) is our framework of MG-FTE under the $2$-way-$1$-shot setting. The color red, blue, and black denote subjects, objects, and other words, respectively. The color orange and green represent different relations. Subfigure (b) visualizes the relationships between entity prototypes and query tokens in the proto-level fusion module.
  }\label{f2}
\vspace{-2mm}
\end{figure*}

\section{Methodology}

\subsection{Problem Definition}
Following the classical few-shot task setting, we define the few-shot triple extraction task in $N$-\textit{way} and $K$-\textit{shot} as well. Each task consists of $K$ labeled sentences per relation, with $N$ relations in the support set,
while the aim is to extract the valid triple $(s, r, o)$ from new sentences in the query set under the same $N$ relations. 
During the training stage, the few-shot extraction task is conducted on a training dataset $\mathcal{D}_{train}$, consisting of a number of labeled instances and relation classes. 
When testing, the task is conducted on the test dataset $\mathcal{D}_{test}$ targeting new relation classes. Note that $\mathcal{D}_{train} \cap \mathcal{D}_{test} = \varnothing$. 
It is worth noting that we also conduct the few-shot triple extraction in a cross-domain manner, where the dataset $\mathcal{D}_{train}$ and $\mathcal{D}_{test}$ belong to different domains. 


\subsection{Entity Prototype Learning}\label{subsec:entity_bio}
We first feed the sentences into the BERT model to obtain the low-dimensional token representation for the query sentence $\mathbf{Q} \in \mathbf{R}^{T_q \times d}$ and support sentence $\mathbf{S}_{i, \, k} \in \mathbf{R}^{T_s \times d}$, where $i$ indicates the $i$-th relation, $k$ denotes the $k$-th sample under relation $r$, $T_q$ and $T_s$ denote the length of the query and support sentences, and $d$ refers to the dimension. 

Considering that the entity in triples may involve multiple tokens, we adopt the BIO tagging scheme \cite{BIO-tag} to build an entity tag set $E = \{BS, IS, BO, IO, O\}$, where $BS$ and $BO$ indicate the beginnings of the subject and the object, respectively; $IS$ and $IO$ indicate the insides of the subject and the object, respectively; $O$ denotes other tokens. The number of entity labels $|E|$ equals to $5$. Therefore, the entity prototypes for relation $ i $ and entity tag $ l $ is  $\mathbf{P}_{i, \, l} \in \mathbb{R}^d $, which is obtained by averaging the corresponding $K$ instances in $\mathcal{D}^{sup}$ as:
\begin{equation}\label{eq:entity_prototype}
    \mathbf{P}_{i, \, l} = \frac{1}{K} \sum_k^K \mathbf{S}_{i, \, k} \cdot \mathbbm{1} \{Y_e == l \} \,,
\end{equation}
where $Y_e = \{y_0, y_1, ..., y_{|s|+1}\}$ denotes the entity tags for each token in the support sentence $\mathbf{S}_{i,k}$, 
The final entity prototypes are $\mathbf{P}^E = \{ \mathbf{P}_i;\, i = 1, ..., N\}$, where $\mathbf{P}_i \in \mathbb{R}^{|E| \times d}$.

\subsection{Entity-Guided Relation Proto-Decoder}\label{subsec:relation_decoder}
Considering that the discrepancy of entities under different relations is large, it would be better to extract relational entities based on the extracted relations. 
However, when relational entities are not available for a query sentence, plenty of non-relational entities and words may mislead the relation classification. To eliminate the negative influence caused by non-relational entities and words, we leverage the entity prototypes learned in Section \ref{subsec:entity_bio} to represent the corresponding relational semantics. In this way, those non-relational entities and words are attributed to one common prototype ``$O$'' and tokens in a query sentence can focus more on the semantics related to a certain relation.

As shown in the Fig.\ref{f2} (a), besides the token representation in $\mathbf{Q}$ and the entity prototypes $\mathbf{P}_i$ under relation $i$, we also
leverage our proto-level fusion module, which will be illustrated in Section \ref{subsec:proto-level_fusion}, to have the fused token representation $\widehat{\mathbf{Q}}_i \in \mathbf{R}^{T_q \times d}$ with the same dimension as $\mathbf{Q}$, and fused entity prototypes $\widehat{\mathbf{P}}_i \in \mathbf{R}^{|E| \times d}$ with the same dimension as $\mathbf{P}_i$. Later, we employ the max-pooling and average-pooling to capture the global semantic representations for each query instance and relation prototype:

\begin{equation}
\begin{aligned}
    & \widetilde{\mathbf{P}}_i = [\max(\widehat{\mathbf{P}}_i); \, \mathrm{avg}(\widehat{\mathbf{P}}_i)] \, , \\
    & \widetilde{\mathbf{Q}}_i = [\max(\widehat{\mathbf{Q}}_i); \, \mathrm{avg}(\widehat{\mathbf{Q}}_i)] \,.
\end{aligned}
\end{equation}

Then, under the relation $i$, we leverage a multi-layer perception (MLP) to measure the matching score between each query instance and the $i$-th relation prototypes as follows:
\begin{equation}
    ms(\widetilde{\mathbf{Q}}_i, \, \widetilde{\mathbf{P}}_i) = \mathbf{V}_r^\intercal \left( \sigma \big( [\widetilde{\mathbf{Q}}_i; \,  \widetilde{\mathbf{P}}_i] \cdot \mathbf{W}_r \big) \right) \, ,
\label{eq:relation_score}
\end{equation}
where $\mathbf{W}_r \in \mathbb{R}^{4d \times d}$ and $\mathbf{V}_r \in \mathbb{R}^{d \times 1}$ are trainable parameters, $\sigma$ denotes the $\mathrm{ReLU}$ activation function. After obtaining the matching score under each relation, we select the relational index with the largest matching score as the classified relation for query instance $\mathbf{Q}$. During training, the cross-entropy loss $\mathcal{L}_{rel} $ is leveraged to calculate the discrepancy between the predicted relation score and the ground truth. 

\subsection{Relation-Guided Entity Proto-Decoder}
Based on the relation classification method in Section \ref{subsec:relation_decoder}, we have known what kind of relational semantics that a sentence in $\mathcal{D}^{qry}$ has. 
Assuming that the query instance $\mathbf{Q}$ belongs to relation $i$, only the entity prototypes and the token representation under relation $i$ are required to calculate the proximity between each token and $|E|$ entity prototypes. This is regarded as the relation-guided entity proto-decoder thanks to the use of the relation classification outcome. 

Instead of simply leveraging the fused token representation $\widehat{\mathbf{Q}}$ for the query instance and corresponding fused entity prototypes $\widehat{\mathbf{P}}_i$ obtained through the proto-level fusion module in Section \ref{subsec:proto-level_fusion}, we also utilize the original token representation  $\mathbf{Q}$ and entity prototypes $\mathbf{P}^E$ for entity extraction. To be specific, for the representation of tokens and entity prototypes, we first concatenate the original and fused features of them to get the representation of tokens $\overline{\mathbf{Q}}_i = \mathrm{concat}([\mathbf{Q}; \widehat{\mathbf{Q}}_i]) \,$ and entity prototypes $ \overline{\mathbf{P}}_i = \mathrm{concat}([\mathbf{P}_i; \widehat{\mathbf{P}}_i]) \,$, respectively.
Then, we use Euclidean distance $ed(x, y) = ||x - y||^2$ to calculate the distances between each token in $\overline{\mathbf{Q}}_i$ and each entity prototype in $\overline{\mathbf{P}}_i$. Note that the operation of concatenating the original and fused representation is to prevent the model from forgetting the characteristics of the entity itself. 
Moreover, we also adopt Conditional Random Field (CRF) \cite{CRF}, consisting of the emission score and transition score, to measure the proximity among different entity labels. The overall entity loss is:

\begin{equation}\label{eq:loss_entity}
    \mathcal{L}_{ent} = \frac{ \mathrm{e} ^ {\left(f_{E}(y \, | \, \overline{\mathbf{Q}}_i, \, \overline{\mathbf{P}}_i) + f_{T}(y) \right)}}{\sum_{y'}^{Y'} \mathrm{e} ^ {\left( f_{E}(y'\, | \, \overline{\mathbf{Q}}_i, \, \overline{\mathbf{P}}_i) + f_{T}(y') \right)}} \,,
\end{equation}
where the emission score $f_{E}(y \, | \, \overline{\mathbf{Q}}_i, \, \overline{\mathbf{P}}_i) = -ed(\overline{\mathbf{Q}}_i, \overline{\mathbf{P}}_i)$, and the transition score $f_{T}(y) = \sum_{j}^{T_q} p(y_j, \,  y_{j+1})$. $y = \{y_1, y_2, ..., y_{T_q}\}$ denotes the entity labels for each token in 
$\mathbf{Q}$, and $Y'$ represents all the possible sequence labels.

Besides, the model utilizes the true relation label of query instances in $\mathcal{D}^{qry}$ to correctly learn the connection between token representations and entity prototypes in the training phase. The predicted relations obtained from the entity-guided relation proto-decoder are leveraged in the test phase. 

\subsection{Proto-level Fusion Module}\label{subsec:proto-level_fusion}
To strengthen the connection between the entity and relation, we design a proto-level fusion module to learn the mixed features of entity prototypes in support set and tokens in query set, which can be leveraged both in entity-guided relation proto-decoder and relation-guided entity proto-decoder. 
In particular, we measure the relationship between each token in $\mathbf{Q}$ and the entity prototypes $\mathbf{P}_i$ under relation $i$ from proto-level, to have the corresponding representations $\mathbf{P}'_i$ and $\mathbf{Q}'_i$:
\begin{equation}
\begin{aligned}
    & \alpha_{i} = \mathbf{Q} \cdot \mathbf{P}_i^\intercal \,, \\
    & \mathbf{P}'_i = \mathrm{softmax}(\alpha_i, \, 1)^\intercal \cdot \mathbf{Q}_i \,, \\
    & \mathbf{Q}'_i = \mathrm{softmax}(\alpha_i, \, 0) \cdot \mathbf{P}_i^\intercal \,,
\end{aligned}
\end{equation}
where $\mathrm{softmax}(x, \, 0)$ and $\mathrm{softmax}(x, \, 1)$ indicate the softmax operations performed by row and column, respectively, and $\cdot$ denotes the dot product. 
Then, a ReLU layer is leveraged to fuse the original and mixed representations as follows:
\begin{equation}\label{eq:entity_match}
\begin{aligned}
    & \widehat{\mathbf{P}}_i = \sigma \big([\mathbf{P}_i; \,  \mathbf{P}'_i; \, |\mathbf{P}_i-\mathbf{P}'_i|; \,  \mathbf{P}_i \odot \mathbf{P}'_i] \cdot \mathbf{W} \big) \,, \\
    & \widehat{\mathbf{Q}}_i = \sigma \big( [\mathbf{Q}_i; \, \mathbf{Q}'_i; \, |\mathbf{Q}_i - \mathbf{Q}'_i|; \, \mathbf{Q}_i \odot \mathbf{Q}'_i] \cdot \mathbf{W} \big) \,,
\end{aligned}    
\end{equation}
where $\odot$ and $[;]$ indicate element-wise product and concatenation, $\mathbf{W} \in \mathbb{R}^{4d \times d}$ are trainable parameters. Finally, the fused representation of the query instance and entity prototypes under each relation are obtained. 
Note that our proto-level fusion module aims at learning the semantics similarity between a query sentence and corresponding relation distribution. Hence, our model can generalize well in another domain and the results in Table \ref{tab:results_cross_domain} prove the efficiency of our method.

\section{Experiments}
\begin{table}[t]\small
\caption{Main results on FewRel 1.0 (F1 score). $\uparrow$ indicates the improvements of our method, while $\downarrow$ illustrates the decline of our framework by removing corresponding parts.}
\label{tab:results_single_domain}
\centering
\begin{tabular}{lcccccc}
\hline 
\specialrule{0em}{0pt}{1pt}
\multicolumn{1}{l}{\textbf{Methods}} &
\multicolumn{1}{c}{\textbf{$5$-way, $5$-shot}} &
\multicolumn{1}{c}{\textbf{$10$-way, $10$-shot}} \\
\hline 
\specialrule{0em}{0pt}{1pt}
BERT \cite{BERT}  &4.71  &2.94\\
CasRel \cite{CasRel}  &2.11  &2.04\\
\hline
\specialrule{0em}{0pt}{1pt}
Proto \cite{prototype}   & 18.35 & 16.92\\
MLMAN \cite{MLMAN}  &19.60 & 17.8\\
MPE \cite{COLING20}   &23.34  &12.08\\
StructShot \cite{structshot} &25.94 &20.28 \\
PA-CRF \cite{pacrf}   &34.14 &30.44\\
RelATE \cite{sigir22}   &42.32 & 40.93\\
\hline 
\specialrule{0em}{0pt}{1pt}
\textbf{MG-FTE}    &\textbf{55.17 ($\uparrow12.85$)} &\textbf{53.33 ($\uparrow12.40$)} \\
\hline
\specialrule{0em}{0pt}{1pt}
 - EG$_r$  &51.27 ($\downarrow3.90$)  &48.08 ($\downarrow5.25$)\\
 - RG$_e$  &44.60 ($\downarrow10.57$)  &33.90 ($\downarrow19.43$) \\
 - PFM   &50.47 ($\downarrow4.70$)  &46.68 ($\downarrow6.65$)\\
\hline 
\end{tabular}
\vspace{-2mm}
\end{table}

\subsection{Experimental Setups}
\textbf{Datasets and Evaluation Metrics.}
We evaluate our model on two widely used few-shot datasets, FewRel 1.0 \cite{FewRel1} for the single domain setting and FewRel 2.0 \cite{FewRel2} for the cross-domain setting. 
Consistent with the MPE \cite{COLING20}, our experiments investigate two few-shot learning configurations, $5$-way-$5$-shot and $10$-way-$10$-shot, by randomly sampling 1,000 tasks from the test set. For the cross-domain setting, we test the model under $3$-way-$3$-shot and $5$-way-$5$-shot settings. We use the micro F1 score to measure the performance of triple extraction.\\
\textbf{Beselines}. 
We divide the compared baseline methods into two categories: supervised learning methods and few-shot learning methods. For supervised learning methods, we finetune the conventional BERT \cite{BERT} and CasRel \cite{CasRel} models on the support set with 10 iterations and test on the query set for each task. 
For few-shot learning methods, we utilize Prototypical Network (Proto) \cite{prototype}, MLMAN \cite{MLMAN}, MPE \cite{COLING20}, StructShot \cite{structshot}, PA-CRF \cite{pacrf} and RelATE \cite{sigir22} as baselines. Besides, for cross-domain triple extraction, we apply two cross-domain strategies, adversarial loss and target-domain pretrained model \cite{RoBERTa-BioMed} to Proto as our cross-domain baselines.

\subsection{Experimental Results}
\textbf{For Single Domain}. 
From Table \ref{tab:results_single_domain}, we have several observations. 1) Our method MG-FTE achieves the best performance compared with all baseline methods. We have 30.36\% and 30.30\% improvements regarding the F1 score for the triple extraction under $5$-way-$5$-shot and $10$-way-$10$-shot settings, respectively. 
3) For ``- EG$_r$'', we fuse the query instances with the token representation of each sentences in support set $\mathcal{D}^{sup}$ instead of the entity prototypes under each relation for entity-guided relation proto-decoder.
The results show that the negative influence caused by non-relational entities and words reduces the performance by 8.46\%.
4) For ``- RG$_e$'', we remove the relation-guided entity proto-decoder, that is, extracting entities under $N$ relations for each task. The performance of triple extraction drops 19.16\% and 36.43\% for $5$-way-$5$-shot and $10$-way-$10$-shot settings on F1 score, which shows the effectiveness of considering entity extraction under specific relations.
5) For ``- PFM'', we delete the proto-level fusion module.
The results show that this module can exactly strengthen the connection between the relational entities and relation extraction with 11.5\% improvements on final results. Besides, we also visualize the output of our PFM in the Fig.\ref{f2} (b), which illustrates that the features of subject ``Jalisse'' are closer to ``BO'' and ``IO'' prototypes, while the features of object ``Fiumi di parole'' are closer to ``BS'' and ``IS'' prototypes under the relation ``performer''. \\ 
\textbf{For Cross Domain}. 
From Table \ref{tab:results_cross_domain}, we observe that
1) our framework achieves better performance in the cross-domain setting with 16x more of the F1 score for relational triple extraction. 
2) ``+ A'' results show that the adversarial loss can deal with the domain adaptation to a certain extent for Proto. Whereas, in our method, we aim at learning the semantics similarity between query sentences and support sentences instead of the semantic representations for each sentence. Therefore, although without adversarial loss, our method can capture the connection between query and support sentences regardless of domains with 27.22\% improvements on the relation classification.
3) The results in the bottom two lines show that the model, pre-trained in the target domain, can better capture domain-invariant features and improve the performance compared with adversarial loss. The improvement mainly lies in the relation classification with about 7.2 F1 score. For entity extraction, the results only show 2.3 improvements on F1 score. A possible reason is that learned domain-invariant features are mainly from the semantic level instead of the entity level, since entities in different domains are distinct. 


\begin{table}[t]\small
\caption{Main results in the cross-domain setting on FewRel 2.0 (F1 score). ``R'' and ``A'' indicate the pre-trained model RoBERTa-BioMed \cite{RoBERTa-BioMed} and adversarial loss, respectively.}
\label{tab:results_cross_domain}
\centering
\begin{tabular}{p{1.2cm}cccccc}
\hline 
\specialrule{0em}{0pt}{1pt}
\multirow{2}{*}{\textbf{Methods}} &
\multicolumn{3}{c}{\textbf{3-way, 3-shot}} &
\multicolumn{3}{c}{\textbf{5-way, 5-shot}} \\
&Entity &Rel &Triple &Entity &Rel &Triple\\
\hline 
\specialrule{0em}{0pt}{1pt}
Proto    &1.39  &65.62 &0.81 &1.54  &54.57  &0.78\\
 + A    &1.76  &66.30 &1.03 &1.97  &55.50  &0.96 \\
 + R    &1.50  &72.81  &1.06 &1.47  &64.29  &0.96 \\
 + R + A    &1.47  &77.07 &1.20 &1.59  &68.89  &1.32 \\
\hline
\specialrule{0em}{0pt}{1pt}
MG-FTE &17.18  &81.22  &17.11  &21.91  &73.22  &21.77   \\  
\textbf{+ R }   &\textbf{18.80}  &\textbf{86.71}  &\textbf{18.80} &\textbf{24.70}  &\textbf{82.10} &\textbf{24.70}  \\
\hline 
\end{tabular}
\vspace{-2mm}
\end{table}

\vspace{-1mm}
\section{Conclusion}
\vspace{-1mm}
In this paper, we propose a mutually guided few-shot learning framework for few-shot relational triple extraction and engage this task into the cross-domain setting. We consider the task from a novel perspective by first classifying relations using an entity-guided relation proto-decoder and then recognizing entities with a relation-guided entity proto-decoder. To consolidate the connection between entities and relations, a proto-level fusion module is designed. The experimental results illustrate the superiority of our method in both single-domain and cross-domain settings.

\section{Acknowledgements}
This work is supported by the NSFC (No.62171323), the National Key R\&D Program of China (No.2020YFA0711400), the Start-up Grant (No.9610564) and the Strategic Research Grant (No.7005847) of City University of Hong Kong.

\printbibliography


\end{document}